\documentclass[runningheads]{llncs}
\usepackage{graphicx}

\usepackage{tikz}
\usepackage{comment}
\usepackage{amsmath,amssymb} 
\usepackage{color}
\usepackage{multirow}
\usepackage{array}
\usepackage{booktabs}
\usepackage{amsfonts}

\usepackage[caption=false]{subfig}
\usepackage{wrapfig}

\usepackage{pifont}
\newcommand{\cmark}{\ding{51}}%
\newcommand{\xmark}{\ding{55}}%
\usepackage{xcolor}
\usepackage[misc]{ifsym}
\usepackage{etoolbox}
\usepackage[pagebackref=true,breaklinks=true,colorlinks,bookmarks=false]{hyperref}
\usepackage[export]{adjustbox}
\usepackage{orcidlink}

\newcommand\Tstrut{\rule{0pt}{2ex}}         
\newcommand\Bstrut{\rule[-1ex]{0pt}{0pt}}   

\newcommand{\repthanks}[1]{\textsuperscript{\ref{#1}}}
\makeatletter
\patchcmd{\maketitle}
  {\def\thanks}
  {\let\repthanks\repthanksunskip\def\thanks}
  {}{}
\patchcmd{\@maketitle}
  {\def\thanks}
  {\let\repthanks\@gobble\def\thanks}
  {}{}
\newcommand\repthanksunskip[1]{\unskip{}}
\makeatother

\begin{document}

\hbadness=2000000000
\vbadness=2000000000
\hfuzz=100pt

\setlength{\abovedisplayskip}{1pt}
\setlength{\belowdisplayskip}{1pt}
\setlength{\floatsep}{6pt plus 1.0pt minus 1.0pt}
\setlength{\intextsep}{6pt plus 1.0pt minus 1.0pt}
\setlength{\textfloatsep}{6pt plus 0pt minus 3.0pt}
\setlength{\parskip}{0pt}
\setlength{\abovedisplayshortskip}{0pt}
\setlength{\belowdisplayshortskip}{0pt}

\pagestyle{headings}
\mainmatter
\def\ECCVSubNumber{5619}  

\title{
Source-free Video Domain Adaptation by Learning Temporal Consistency for\\ Action Recognition\thanks{This research is jointly supported by A*STAR Singapore under its AME Programmatic Funds (Grant No.\ A20H6b0151) and Career Development Award (Grant No. C210112046), and by Nanyang Technological University, Singapore, under its NTU Presidential Postdoctoral Fellowship, ``Adaptive Multimodal Learning for Robust Sensing and Recognition in Smart Cities'' project fund.}
}


\titlerunning{SFVDA by Learning Temporal Consistency}
\author{Yuecong Xu\inst{1}\thanks{Equal Contributions\protect\label{X}}\orcidlink{0000-0002-4292-7379}\index{Xu, Yuecong} \and 
Jianfei Yang\inst{2}\repthanks{X}\orcidlink{0000-0002-8075-0439}\index{Yang, Jianfei} \and 
Haozhi Cao\inst{2}\orcidlink{0000-0001-7703-3490}\index{Cao, Haozhi} \and \\ 
Keyu Wu\inst{1}\orcidlink{0000-0001-8493-0712}\index{Wu, Keyu} \and 
Min Wu\inst{1}\orcidlink{0000-0003-0977-3600}\index{Wu, Min} \and 
Zhenghua Chen\inst{1}(\Letter)\orcidlink{0000-0002-1719-0328}}\index{Chen, Zhenghua} 
\authorrunning{Y. Xu, J. Yang, H. Cao, K. Wu, M. Wu, and Z. Chen}
\institute{
Institute for Infocomm Research, A*STAR, Singapore\\
\email{xuyu0014@e.ntu.edu.sg, \{wu\textunderscore keyu, wumin\}@i2r.a-star.edu.sg, chen0832@e.ntu.edu.sg}\and
School of Electrical and Electronic Engineering,\\ Nanyang Technological University, Singapore\\
\email{\{yang0478,haozhi001\}@ntu.edu.sg}\\
}

\maketitle

\begin{abstract}
    Video-based Unsupervised Domain Adaptation (VUDA) methods improve the robustness of video models, enabling them to be applied to action recognition tasks across different environments. However, these methods require constant access to source data during the adaptation process. Yet in many real-world applications, subjects and scenes in the source video domain should be irrelevant to those in the target video domain. With the increasing emphasis on data privacy, such methods that require source data access would raise serious privacy issues. Therefore, to cope with such concern, a more practical domain adaptation scenario is formulated as the \textit{Source-Free Video-based Domain Adaptation} (SFVDA). Though there are a few methods for Source-Free Domain Adaptation (SFDA) on image data, these methods yield degenerating performance in SFVDA due to the multi-modality nature of videos, with the existence of additional temporal features. In this paper, we propose a novel Attentive Temporal Consistent Network (ATCoN) to address SFVDA by learning temporal consistency, guaranteed by two novel consistency objectives, namely feature consistency and source prediction consistency, performed across local temporal features. ATCoN further constructs effective overall temporal features by attending to local temporal features based on prediction confidence. Empirical results demonstrate the state-of-the-art performance of ATCoN across various cross-domain action recognition benchmarks. Code is provided at \href{https://github.com/xuyu0010/ATCoN}{https://github.com/xuyu0010/ATCoN}.
    
\end{abstract}

\section{Introduction}
\label{section:intro}

Video-based tasks such as action recognition have long been investigated considering their wide applications. Deep neural networks have made remarkable advances with the introduction of large-scale labeled datasets~\cite{kay2017kinetics,monfort2019moments}. However, due to the expense of laborious video data annotation, sufficient labeled training videos may not be readily available in real-world scenarios. To avoid costly data annotation, various \textit{Video-based Unsupervised Domain Adaptation} (VUDA) methods have been introduced to transfer knowledge from a labeled source video domain to an unlabeled target video domain by reducing discrepancies between source and target video domains~\cite{chen2019temporal,choi2020shuffle,xu2021multi}. VUDA methods greatly improve the robustness of video models, enabling them to be applied to action recognition tasks across different environments~\cite{xu2021aligning}.

Though current VUDA methods~\cite{chen2019temporal,choi2020shuffle,xu2021partial,xu2021aligning} enable the transfer of knowledge across video domains, they all require access to source video data during the adaptation process. Yet action information usually contains the private and sensitive information of the actors, including their actions and the relevant scenes. Meanwhile, in real-world applications, such information in the source domain is usually irrelevant to those in the target domain and should be protected from the target domain. Therefore, current VUDA methods would raise serious privacy issues, which is more severe than that raised by image-based domain adaptation. To cope with the video data privacy issue, a more practical domain adaptation scenario is formulated as the \textit{Source-Free Video-based Domain Adaptation} (SFVDA), where only well-trained source video models and unlabeled target domain data would be provided for adaptation. 

With the absence of source data, current VUDA methods that mainly align target and source domains statistically~\cite{long2015learning,sun2016return} cannot be applied to the SFVDA problem. Recently, there are a few research efforts~\cite{li2020model,liang2021source,yeh2021sofa} that start exploring Source-Free Domain Adaptation (SFDA) with image data, where SFDA is tackled by adjusting target features to adapt to the source classifier~\cite{liang2020we}. The key idea is to learn discriminative latent target features while aligning source data distribution embedded within the source classifier. However, aligning videos without source data is even more challenging thanks to the fact that videos are characterized by their multi-modality nature, where temporal features are key components that are excluded in images. 

While direct minimization of statistical discrepancy between target and source domains cannot be achieved due to the lack of source data, domain adaptation can also be achieved by aligning the embedded semantic information~\cite{xie2018learning,li2021semantic} via entropy-based approaches~\cite{saito2019semi,vu2019advent} such as maximizing mutual information~\cite{viola1997alignment} or neighborhood clustering~\cite{saito2020universal}. These methods improve the discriminability of the target features which satisfy the cluster assumption~\cite{grandvalet2004semi}, while increasing the source model transferability~\cite{yang2020mind}. However, these methods are insufficient for aligning semantic information in videos. The reason is that overall temporal feature of a video can be constructed with a series of local temporal features, obtained through clips sampled from videos. Each local temporal feature should be discriminative in the first place. However, if each local temporal feature is individually discriminative yet mutually inconsistent, the local temporal features may not hold similar semantic information. Subsequently, the overall temporal feature may contain indistinct semantic information, and would not be discriminative. Instead, we hypothesize that for source videos, the extracted local temporal features are not only discriminative, but also consistent across each other and possess similar feature distribution patterns, which implies similar semantic information. Such hypothesis is termed as the \textit{cross-temporal hypothesis}. If the target data aligns with the source data distribution, we assume that source-like representations are learned for target data, therefore the \textit{cross-temporal hypothesis} should be satisfied by the target data representation. To this end, our method is designed such that the local temporal features are consistent in their feature representations, which would result in the corresponding overall temporal feature being effective and discriminative.

Meanwhile, since only the source model with the source classifier is available for adaptation, the relevance of the target data to source data distribution is highly correlated to the prediction of target data on the source classifier. Therefore, to better adapt target temporal features to the source classifier, the relevance of the corresponding local temporal features towards source data distribution should also be consistent. Such consistency can be interpreted as the source prediction consistency of local temporal features with respect to the fixed source classifier. Further, to improve the discriminability of the video feature, the overall temporal feature should be built by an attentive combination of local temporal features. The attentive combination builds upon the confidence of each local temporal feature towards its relevance to source data distribution. 

To this end, we propose an \textbf{Attentive Temporal Consistent Network (ATCoN)} to address SFVDA uniformly. ATCoN leverages temporal features effectively by learning \textbf{temporal consistency} via \textbf{feature consistency} and \textbf{source prediction consistency} for local temporal features in a self-supervised manner. ATCoN further adapts target data to the source data distribution by attending to local temporal features with higher confidence over its relevance towards source data distribution, indicated as higher source prediction confidence.

In summary, our contributions are threefold. First, we formulated a practical and challenging \textit{Source-Free Video Domain Adaptation} (SFVDA) problem. To the best of our knowledge, this is the first research that studies source-free transfer for video-based tasks, which aims to address data-privacy issues in VUDA. Secondly, we analyze the challenges underlying SFVDA and propose ATCoN to address the challenges uniformly. ATCoN aims to obtain effective and discriminative overall temporal features that satisfies the \textit{cross-temporal hypothesis} by learning temporal consistency which is composed of both feature and source prediction consistency. ATCoN further aligns target data to the source data distribution without source data access by attending to local temporal features with high source prediction confidence. Finally, empirical results demonstrate the efficacy of our proposed ATCoN, achieving state-of-the-art performance across the multiple cross-domain action recognition benchmarks.

\section{Related Work}
\label{section:related}

\textbf{Unsupervised Domain Adaptation (UDA) and Video-based Unsupervised Domain Adaptation (VUDA).} 
Current UDA and VUDA methods aim to distill shared knowledge across labeled source domains and unlabeled target domains. These methods improve the transferability and robustness of models. Generally, they could be divided into three categories: a) reconstruction-based methods~\cite{ghifary2016deep,yang2020label}, where domain-invariant features are obtained by encoders trained with data-reconstruction objectives, whose methods are commonly formulated as encoder-decoder networks; b) adversarial-based methods~\cite{chen2019temporal,xu2021aligning}, where domain-invariant features are extracted by feature generators while leveraging domain discriminators, which are trained jointly in an adversarial manner~\cite{huang2011adversarial}, minimizing adversarial losses~\cite{ganin2015unsupervised}; and c) discrepancy-based methods \cite{saito2018maximum,zhang2019bridging,yang2021advancing}, which mitigate domain shifts across domains by applying metric learning approaches, minimizing metrics such as MMD~\cite{long2015learning} and CORAL~\cite{sun2016return}. By comparison, VUDA research lags behind UDA research, mainly due to the challenges brought by aligning temporal features in videos. However, with the introduction of various cross-domain video datasets such as UCF-HMDB\textsubscript{\textit{full}}~\cite{chen2019temporal} and Sports-DA~\cite{xu2021multi}, there has been a significant increase in research interests for VUDA~\cite{choi2020shuffle,pan2020adversarial,chen2020action}. Despite the improvements in video model robustness brought by VUDA methods, all such methods require access to source data during the adaptation process. Such requirements could raise serious privacy concerns given the amount of private information of the relevant subject and scene in videos.

\noindent\textbf{Source-Free Domain Adaptation (SFDA).}
With the increased importance of data privacy, there have been a few recent research efforts that investigate SFDA with images, which enable image models to be adapted to the target domain without access to source data. Among them, 3C-GAN~\cite{li2020model} and SDDA~\cite{kurmi2021domain} seek to produce novel target-style data that are similar to the source domain. Domain invariant features are then obtained by aligning the novel target-style data with the original target data via adversarial-based domain adaptation methods. Similarly, CPGA~\cite{qiu2021cpga} tackles SFDA by generating avatar feature prototypes for each class, which are trained with the target features in an adversarial manner. Meanwhile, SHOT~\cite{liang2021source,liang2020we} exploits knowledge of source feature distribution by freezing the source classifier and matches target features to the source classifier by leveraging information maximization and pseudo-labeling. More recently, BAIT~\cite{yang2020bait} extends MCD~\cite{saito2018maximum} to SFDA. Despite the advances made in the research of SFDA for images, SFVDA has not been tackled. Due to the amount of private data in videos, SFVDA is even more critical, yet is also more challenging given that temporal features must also be aligned. We propose to engage in SFVDA by utilizing temporal features via learning temporal consistency while attending to local temporal features with high confidence.

\section{Proposed Method}
\label{section:method}

In the scenario of \textit{Source-Free Video Domain Adaptation} (SFVDA), we are only given a source video model that consists of the spatial feature extractor $G_{S,sp}$, the temporal feature extractor $G_{S,t}$ and the classifier $H_{S}$, and an unlabeled target domain $\mathcal{D}_{T}=\{V_{iT}\}^{n_{T}}_{i=1}$ with $n_{T}$ i.i.d.\ videos, characterized by a probability distribution of $p_{T}$. The source model is generated by training its parameters $\theta_{S,sp}$, $\theta_{S,t}$, and $\theta_{H}$ with the labeled source domain $\mathcal{D}_S=\{(V_{iS},y_{iS})\}^{n_S}_{i=1}$ containing $n_{S}$ videos. We assume that both the labeled source domain videos and the unlabeled target domain videos share the same $C$ classes, yet $\mathcal{D}_S$ is inaccessible when adapting the source model to $\mathcal{D}_{T}$.

Owing to the absence of the source domain during adaptation, SFVDA is more challenging while current VUDA methods cannot be applied. SFVDA should be tackled by adapting target video features to the source classifier, which contains information regarding source data distribution. The core is to extract source-like representations that satisfy the \textit{cross-temporal hypothesis}, characterized by the consistency across local temporal features. We propose ATCoN, a novel network to transfer source models to the target domain by leveraging temporal features constructed attentively through learning temporal consistency in a self-supervised manner. We start with an introduction to the generation of the source model, followed by a thorough illustration of ATCoN.

\subsection{Source Model Generation}
\label{section:method:src-model}
A key prior for the transferred model to obtain effective temporal features is that the generated source model could extract precise temporal features. While conventional 3D-CNN-based extractors (e.g., 3D-ResNet~\cite{hara2017learning} or I3D~\cite{carreira2017quo}) have been adopted in action recognition due to their performances, they extract spatio-temporal features jointly while temporal features are obtained implicitly by temporal pooling. In contrast, the Temporal Relation Network (TRN)~\cite{zhou2018temporal} is adopted for SFVDA, thanks to its ability in obtaining more precise temporal features through reasoning over correlations between spatial representations, which corresponds with how humans would recognize actions.

Formally, an input source video with $k$ frames can be expressed as $V_{iS}=\{f_{iS}^{(1)},f_{iS}^{(2)},...,f_{iS}^{(k)}\}$, where $f_{iS}^{(j)}$ is the spatial representation of the $j-$th frame in the $i-$th source video obtained from the source spatial feature extractor $G_{S,sp}$. $G_{S,sp}$ is formulated as a 2D-CNN (e.g., ResNet~\cite{he2016deep}). The temporal feature of $V_{iS}$ is subsequently obtained from the source temporal feature extractor $G_{S,t}$, constructed by a combination of multiple local temporal features. Each local temporal feature is built upon clips with $r$ temporal-ordered sampled frames where $r\in [2,k]$. Formally, a local temporal feature for $V_{iS}$, $lt_{iS}^{(r)}$, is defined by:
\begin{equation}
\label{eqn:method:loc-t-src}
lt_{iS}^{(r)} = \sum\nolimits_{m} g_{S}^{(r)}((V_{iS}^{(r)})_m),
\end{equation}
where $(V_{iS}^{(r)})_m=\{f_{iS}^{(a)},f_{iS}^{(b)},...\}_m$ is the $m-$th clip with $r$ temporal-ordered frames. $a$ and $b$ are the frame indices, which may not be consecutive as the clip with temporal-ordered frames could be extracted with nonconsecutive frames, but should be both in the range of $[1,k]$ with $b>a$. The local temporal feature $lt_{iS}^{(r)}$ is computed by fusing the time ordered frame-level spatial features through the integration function $g_{S}^{(r)}$, implemented as a Multi-Layer Perceptron (MLP). $G_{S,t}$ is therefore a set of all integration functions $g_{S}^{(r)}$, namely $G_{S,t}=\{\forall_r g_{S}^{(r)}\}$. The final overall temporal feature $\mathbf{t}_{iS}$ is a simple mean aggregation applied across all local temporal features, defined as:
\begin{math}
\mathbf{t}_{iS} = \frac{1}{k-1} \sum\nolimits_{r} lt_{iS}^{(r)}.
\end{math}
The source prediction is further computed by applying a source classifier $H_{S}$ over $\mathbf{t}_{iS}$. The source model is trained with the standard cross-entropy loss as the objective function, formulated as:
\begin{equation}
\label{eqn:method:ce-src}
\mathcal{L}_{S,ce} = -\frac{1}{n_{S}}\sum\nolimits_{i=1}^{n_{S}} y_{iS}\log \sigma (H_{S}(\mathbf{t}_{iS})),
\end{equation}
where $\sigma$ is the softmax function whose $c$-th element is defined as $\sigma_{c}(x) = exp(x_{c})\,/\,\sum\nolimits_{c=1}^{C}exp(x_{c})$. Inspired by~\cite{liang2021source}, for the source model to be more discriminative and transferrable for better target data alignment, we further adopt the label smoothing technique~\cite{szegedy2016smoothing} such that extracted features are encouraged to be distributed in tight clusters evenly separated~\cite{muller2019label}. By adopting the label smoothing technique, the objective function for training the source model can be further formulated as:
\begin{equation}
\label{eqn:method:lsce-src}
\mathcal{L}_{S,ce}^{\prime} = -\frac{1}{n_{S}}\sum\nolimits_{i=1}^{n_{S}} y_{iS}^{\prime}\log \sigma (H_{S}(\mathbf{t}_{iS})),
\end{equation}
where $y_{iS}^{\prime}$ is the smoothed label computed as $y_{iS}^{\prime} = (1-\epsilon)y_{iS} + \epsilon\,/\,C$ with $\epsilon$ being the smoothing parameter which is set to 0.1 empirically.

\begin{figure}[!t]
\begin{center}
   \includegraphics[width=.85\linewidth]{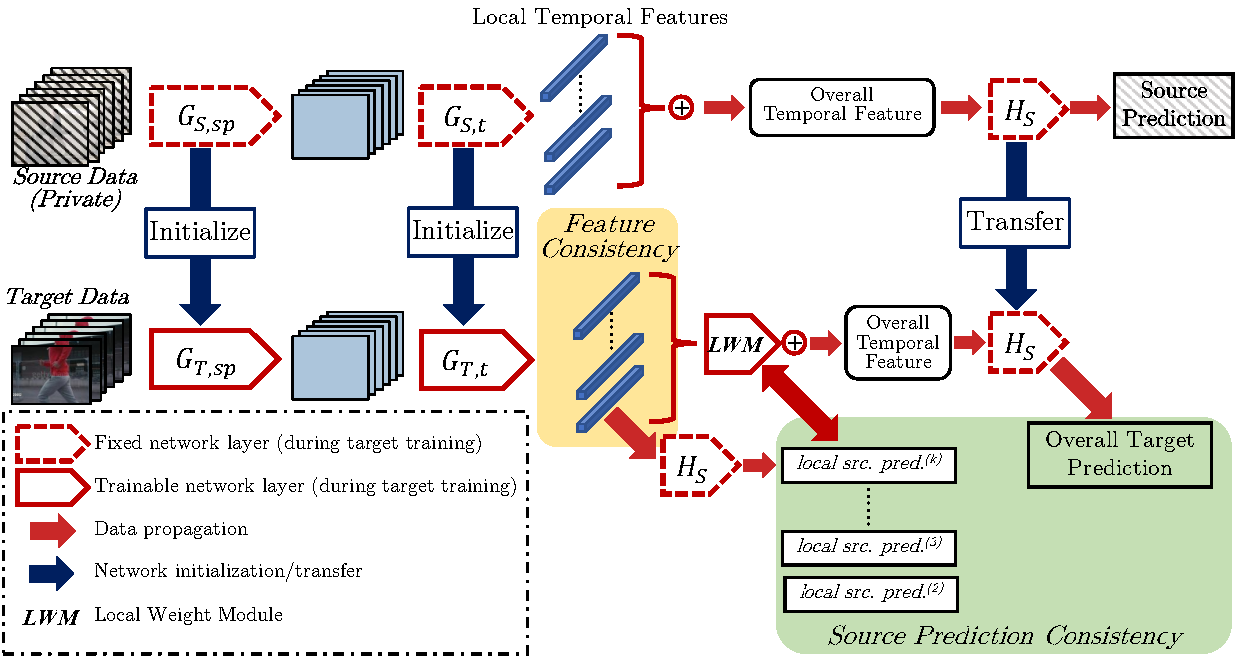}
\end{center}
\caption{Structure of the proposed ATCoN. ATCoN adopts the same network architecture for its spatial and temporal feature extractors as the source model, initialized by the source feature extractors. ATCoN extracts overall temporal features by learning \textit{temporal consistency} over its local temporal features which includes both \textit{feature consistency} and \textit{source prediction consistency}. The \textit{local weight module (LWM)} attends to more confident local temporal features. The overall target prediction is obtained by applying the \textit{fixed} source classifier over the overall temporal feature. Dashed shapes indicates fixed network layers during adaptation.}
\label{figure:3-1-struct}
\end{figure}

\subsection{Attentive Temporal Consistent Network}
\label{section:method:atcon}
With the absence of source data, conventional VUDA methods can no longer be applied. Instead, we tackle SFVDA from two perspectives: on the one hand, extracting effective overall temporal features that are discriminative and comply with the \textit{cross-temporal hypothesis} in a self-supervised manner, without either target label or source data; on the other hand, aligning to the source data distribution via attending to local temporal features with higher confidence in its relevance towards the source data distribution. Following the above strategies, we develop an \textbf{Attentive Temporal Consistent Network (ATCoN)}, whose structure is presented in Fig. \ref{figure:3-1-struct}. With the same network architecture adopted for the target spatial and temporal feature extractors $G_{T,sp}$ $G_{T,t}$ as that of $G_{S,sp}$ $G_{S,t}$, $G_{T,sp}$ and $G_{T,t}$ are initialized by $G_{S,sp}$ and $G_{S,t}$ respectively. The overall temporal feature is obtained by learning temporal consistency over the local temporal features as well as the respective local source predictions, resulted by applying the source classifier $H_{S}$ over the local temporal features directly. Note that the source classifier remains \textit{fixed} throughout the adaptation process. Meanwhile, for attentive aggregation of target local temporal features, a \textit{Local Weight Module (LWM)} is further designed.

\subsubsection{Learning Temporal Consistency.}
\label{section:method:atcon:ltc}
As presented in Section \ref{section:method:src-model}, the different local temporal features are extracted via the multiple temporal-ordered frames, sampled from the input video. For a given input video, these local temporal features should represent the same action even if they may differ in spatial appearances. Therefore, the overall temporal feature is effective and discriminative when the corresponding local temporal features are consistent in feature representations. Given a target input video $V_{T}\in \mathcal{D}_{T}$ (with video index $i$ omitted for simplicity), its local temporal features for the set of clips with $r1$ and $r2$ temporal frames ($r1,r2\in [2,k]$), $lt_{T}^{(r1)}$ and $lt_{T}^{(r2)}$, are defined similarly to Eq. \ref{eqn:method:loc-t-src}. If the local temporal features are consistent, then the cross-correlation matrix between $lt_{T}^{(r1)}$ and $lt_{T}^{(r2)}$ should be close to the identity matrix. The cross-correlation matrix is formulated by:
\begin{equation}
\label{eqn:method:ccm}
\mathcal{C}^{r1r2} = \left(\hat{lt}_{T}^{(r1)}\right)^{T} \hat{lt}_{T}^{(r2)},
\end{equation}
where $\hat{lt}$ is the normalized local temporal feature computed as:
\begin{equation}
\label{eqn:method:norm-lt}
\hat{lt} = \frac{lt - \mathbb{E}(lt)}{\sqrt{Var(lt) + \varepsilon}},
\end{equation}
with $\varepsilon$ being a small bias value for numerical stability. The cross-correlation matrix $\mathcal{C}^{r1r2}$ is a square matrix with the size of $d\times d$, where $d$ is the dimension of the local temporal feature. Since $\mathcal{C}^{r1r2}$ should ideally be close to an identity matrix, the feature consistency loss should maximize the similarity of the respective local temporal features while reducing redundancy between the components. 
Therefore, the feature consistency loss with respect to $\mathcal{C}^{r1r2}$ is expressed as:
\begin{equation}
\label{eqn:method:feat-con-single}
\resizebox{.6\linewidth}{!}{%
\begin{math}
\mathcal{L}_{fc}^{r1r2} = \sum\nolimits_{i}(1-\mathcal{C}_{ii}^{r1r2})^{2} + \lambda \sum\nolimits_{i} \sum\nolimits_{j\neq i} (\mathcal{C}_{ij}^{r1r2})^{2},
\end{math}
}
\end{equation}
where $i,j\in [0,d-1]$ are indexes of the local temporal feature dimension, while $\lambda$ is a tradeoff constant.
The final feature consistency loss is computed as the mean feature consistency loss over all cross-correlation matrices, with each matrix corresponding to a pair of local temporal features. The final feature consistency loss can be formulated as:
\begin{equation}
\label{eqn:method:feat-con-all}
\resizebox{.45\linewidth}{!}{%
\begin{math}
\mathcal{L}_{fc} = \frac{1}{N_{fc}} \left( \sum\nolimits_{r1} \sum\nolimits_{r2\neq r1} \mathcal{L}_{fc}^{r1r2} \right),
\end{math}
}
\end{equation}
where $N_{fc}=\!P_{2}^{k-1}$ is the total number of local temporal feature pairs.

Moreover, since the local temporal features of the same input video should be consistent by minimizing $\mathcal{L}_{fc}$, their relevance towards the source data distribution should also be consistent. With source data inaccessible, such relevance cannot be computed directly through measuring the divergence between source and target data distributions. Since the source classifier contains source data distribution, such relevance could instead be approximated by the prediction of the source classifier over the local temporal features. In other words, the consistency over the relevance of target local temporal features towards source data distribution is equivalent to the consistency over the source prediction of target local temporal features. Meanwhile, the target overall temporal feature is obtained by aggregating the respective local temporal features. It should contain similar motion information as the local temporal features. Therefore, the consistency over source prediction could be extended to the overall temporal feature.

Given local temporal features $lt_{T}^{(2)}, \ldots, lt_{T}^{(k)}$, the respective local source predictions $p_{lt,T}^{(2)}, \ldots, p_{lt,T}^{(k)}$ are obtained via the fixed source classifier $H_{S}$, following:
$p_{lt,T}^{(r)} = H_{S}(lt_{T}^{(r)}),\: \forall r\in[2,k]$.
An average local source prediction could be obtained by averaging over the local source predictions $\bar{p}_{lt,T} = \frac{1}{k-1} \sum\nolimits_{r=2}^{k} p_{lt,T}^{(r)}$. To achieve source prediction consistency, we aim to minimize the divergence between each local source predictions and the average local source prediction:
\begin{equation}
\label{eqn:method:pred-con-loc}
\resizebox{.65\linewidth}{!}{%
\begin{math}
\mathcal{L}_{pc}^{local} = \frac{1}{k-1} \left( \sum\nolimits_{r=2}^{k} KL(\log\sigma(p_{lt,T}^{(r)}) \| \log\sigma(\bar{p}_{lt,T})) \right),
\end{math}
}
\end{equation}
where $KL(p\|q)$ denotes the Kullback–Leibler (KL) divergence.

Further, the overall target prediction $p_{t,T}$ is computed by applying $H_{S}$ to the target overall temporal feature $\mathbf{t}_{T}$, which is a simple mean aggregation applied across local temporal features $lt_{T}^{(2)}, \ldots, lt_{T}^{(k)}$. To incorporate $p_{t,T}$ into the source prediction consistency, we aim to minimize the absolute difference between $p_{t,T}$ and $\bar{p}_{lt,T}$, defined as:
\begin{equation}
\label{eqn:method:pred-con-overall}
\mathcal{L}_{pc}^{overall} = \sum\nolimits_{c=1}^{C} |\log\sigma_{c}(p_{t,T}) - \log\sigma_{c}(\bar{p}_{lt,T})|.
\end{equation}
The final source prediction consistency is achieved by joint minimization of the prediction divergence between each local source prediction and the average local source prediction, as well as between the overall target prediction and the average local source prediction, formulated as:
$\mathcal{L}_{pc} = \alpha_{local} \mathcal{L}_{pc}^{local} + \alpha_{overall} \mathcal{L}_{pc}^{overall}$,
where $\alpha_{local}$ and $\alpha_{overall}$ are tradeoff constants. Learning temporal consistency is thus achieved by optimizing both the source prediction consistency loss and feature consistency loss jointly, expressed as:
$\mathcal{L}_{tc} = \beta_{fc} \mathcal{L}_{fc} + \beta_{pc} \mathcal{L}_{pc}$,
with $\beta_{fc}$ and $\beta_{pc}$ being the tradeoff hyperparameters.

\subsubsection{Local Weight Module (LWM).}
\label{section:method:atcon:lwm}
While complying with the \textit{cross-temporal hypothesis} via learning temporal consistency with feature and source prediction consistencies enables ATCoN to extract discriminative temporal features, we observe that the overall temporal feature $\mathbf{t}_{T}$ is constructed by simply averaging over all local temporal features. This would not be reasonable as the importance of each local temporal feature is commonly uneven. Therefore, we propose the \textit{Local Weight Module (LWM)} to assign \textit{local weights} to the local temporal features for subsequent attentive aggregation.

As mentioned in Section \ref{section:method:atcon}, ATCoN aims to tackle SFVDA by aligning target videos to the source data distribution. Therefore, LWM is designed such that local temporal features that are more confident towards its relevance to the source data distribution gains more attention, weighted by a \textit{local relevance weight}. More specifically, following Section \ref{section:method:atcon:ltc}, the relevance towards source data distribution for $lt_{T}^{(r)}$ could be referred to its local source prediction $p_{lt,T}^{(r)} = H_{S}(lt_{T}^{(r)})$, from which the confidence score is computed. Subsequently, the confidence of $p_{lt,T}^{(r)}$ is defined as the additive inverse of its entropy computed over probabilities of all classes, formulated as:
\begin{equation}
\label{eqn:method:lt-confidence}
\mathbb{C}(p_{lt,T}^{(r)}) = \sum\nolimits_{c=1}^{\mathcal{C}} \sigma_{c}(p_{lt,T,c}^{(r)})\, \log\sigma_{c}(p_{lt,T,c}^{(r)}).
\end{equation}
The \textit{local relevance weight} corresponding to the local temporal feature $lt_{T}^{(r)}$ is finally generated by adding a residual connection for more stable optimization, expressed as:
\begin{math}
w_{lt_{T}^{(r)}} = 1 + \mathbb{C}(p_{lt,T}^{(r)}).
\end{math}
The \textit{local relevance weight} is applied to obtain the weighted overall temporal feature $\mathbf{t}_{T}^{\prime}$, which is the mean aggregation of the corresponding weighted local temporal features, computed as:
$\mathbf{t}_{T}^{\prime} = \frac{1}{k-1} \sum\nolimits_{r} w_{lt_{T}^{(r)}}\, lt_{T}^{(r)}.$
Meanwhile, \textit{local relevance weight} is further applied to the local source predictions $p_{lt,T}^{(r)}$, where the source prediction consistency is learnt with relevance-weighted local source predictions ${p_{lt,T}^{(r)}}^{\prime}= w_{lt_{T}^{(r)}}p_{lt,T}^{(r)}$.

ATCoN learns temporal consistency by learning feature consistency and source prediction consistency of local temporal features jointly. Inspired by prior works in SFDA~\cite{liang2021source,kim2021domain,xia2021adaptive}, we further improve ATCoN from two aspects: 

\subsubsection{Information Maximization.}
The ideal overall temporal feature should be both individually certain and globally diverse. Therefore, we apply an Information Maximization (IM) loss over the weighted overall temporal feature as:
.\begin{equation}
\label{eqn:method:im-t}
\resizebox{.65\linewidth}{!}{%
\begin{math}
\begin{aligned}
\mathcal{L}_{IM} &= -\mathbb{E}_{V_{T}\in \mathbf{D}_{T}} \sum\nolimits_{c=1}^{C} \sigma_{c}(H_{S}(\mathbf{t}_{T}^{\prime}\left(V_{T}\right)))\, \log\sigma_{c}(H_{S}(\mathbf{t}_{T}^{\prime}\left(V_{T}\right)))\\
&+ \sum\nolimits_{c=1}^{C} KL \left(\mathbb{E}_{V_{T}\in \mathbf{D}_{T}}[\sigma_{c}(H_{S}(\mathbf{t}_{T}^{\prime}\left(V_{T}\right)))]\,\|\,\frac{1}{C} \right),
\end{aligned}
\end{math}
}
\end{equation}
where $\mathbf{t}_{T}^{\prime}\left(V_{T}\right)$ is the weighted overall temporal feature corresponding to target video $V_{T}$, while $\sigma_{c}$ is the $c$-th element in the softmax. 

\subsubsection{Self-supervised Pseudo-label Generation.}
To further improve class-wise alignment of ATCoN with the lack of target label, we follow~\cite{liang2020we} and generate pseudo-labels for target videos in a self-supervised manner. Specifically, pseudo-labels are generated through a repeated k-means clustering process over the overall temporal feature, where the initial centroid for class $c$ is attained by:
\begin{equation}
\label{eqn:method:init-centroid}
\resizebox{.5\linewidth}{!}{%
\begin{math}
\mathbf{c}_{c}^{(0)} = \frac{\sum\nolimits_{V_{T}\in \mathbf{D}_{T}} \sigma_{c}(H_{S}(\mathbf{t}_{T}^{\prime}\left(V_{T}\right)))\:\mathbf{t}_{T}^{\prime}\left(V_{T}\right)} {\sum\nolimits_{V_{T}\in \mathbf{D}_{T}} \sigma_{c}(H_{S}(\mathbf{t}_{T}^{\prime}\left(V_{T}\right)))}.
\end{math}
}
\end{equation}
Subsequently, the initial pseudo-label of target data $V_{T}$ is obtained by its nearest centroid, defined by:
$\hat{y}_{V_{T}} = \arg \min_{c} \cos(\mathbf{t}_{T}^{\prime}\left(V_{T}\right),\, \mathbf{c}_{c}^{(0)})$,
where $\cos(\cdot,\cdot)$ denotes the cosine distance function. The initial centroids are further updated to characterize the category distribution of the target domain more reliably based on the initial pseudo-labels, formulated as:
\begin{equation}
\label{eqn:method:update-centroid}
\resizebox{.4\linewidth}{!}{%
\begin{math}
\mathbf{c}_{c}^{(1)} = \frac{\sum\nolimits_{V_{T}\in \mathbf{D}_{T}}  \mathbb{I}(\hat{y}_{V_{T}}=c)\:\mathbf{t}_{T}^{\prime}\left(V_{T}\right)} {\sum\nolimits_{V_{T}\in \mathbf{D}_{T}} \mathbb{I}(\hat{y}_{V_{T}}=c)},
\end{math}
}
\end{equation}
with $\mathbb{I}(\cdot)$ being an indicator function. The pseudo-labels are finally renewed following the updated centroids with $\hat{y}_{V_{T}} = \arg \min_{c} \cos(\mathbf{t}_{T}^{\prime}\left(V_{T}\right),\, \mathbf{c}_{c}^{(1)})$. ATCoN is further trained with the cross-entropy loss with respect to the pseudo-labels as:
\begin{equation}
\label{eqn:method:ce-tgt}
\mathcal{L}_{T,ce} = -\frac{1}{n_{T}} \sum\nolimits_{i=1}^{n_{T}} \hat{y}_{V_{T} }\log\sigma (H_{S}(\mathbf{t}_{T}^{\prime}\left(V_{T}\right))),
\end{equation}
where $n_{T}$ is the total number of target videos. 

\subsubsection{Overall Objective.}
In summary, given a trained source model, the overall optimization objective of ATCoN is expressed as:
$\mathcal{L} = \beta_{tc}\mathcal{L}_{tc} + \beta_{IM}\mathcal{L}_{IM} + \beta_{ce}\mathcal{L}_{T,ce}$,
where $\beta_{tc}$, $\beta_{IM}$, and $\beta_{ce}$ are tradeoff hyperparameters.

\section{Experiments}
\label{section:exps}

In this section, we evaluate our proposed ATCoN across three cross-domain action recognition benchmarks including UCF-HMDB\textsubscript{\textit{full}}~\cite{chen2019temporal}, Daily-DA~\cite{xu2021multi} and Sports-DA~\cite{xu2021multi}. These benchmarks cover a wide range of cross-domain scenarios. We present superior results on all benchmarks. Further, ablation studies and empirical analysis of ATCoN are also presented to validate the architecture of ATCoN. \textit{Code is provided at \href{https://github.com/xuyu0010/ATCoN}{https://github.com/xuyu0010/ATCoN}.}

\subsection{Experimental Settings}
\label{section:exps:setting}
Among the three benchmarks, \textbf{UCF-HMDB\textsubscript{\textit{full}}} is one of the most widely used cross-domain video dataset, which contains videos from two public datasets: UCF101 (U101)~\cite{soomro2012ucf101} and HMDB51 (H51)~\cite{kuehne2011hmdb}, a total of 3,209 videos in 12 action classes, with 2 cross-domain action recognition tasks. Meanwhile, \textbf{Daily-DA} is a more challenging dataset that incorporates both normal videos and low-illumination videos. It is constructed from four datasets : ARID (A11)~\cite{xu2021arid}, HMDB51 (H51), Moments-in-Time (MIT)~\cite{monfort2019moments}, and Kinetics (K600)~\cite{kay2017kinetics}. While HMDB51, Moments-in-Time, and Kinetics are widely used for action recognition benchmarking, ARID is a more recent dark dataset, comprised with videos shot under adverse illumination conditions. In total, \textbf{Daily-DA} includes 18,949 videos from 8 classes, with a total of 12 cross-domain action recognition tasks. \textbf{Sports-DA} is a large-scale cross-domain video dataset, built from UCF101 (U101), Sports-1M (S1M)~\cite{karpathy2014large}, and Kinetics (K600), with 23 action classes and a total of 40,718 videos. With three different domains, \textbf{Sports-DA} contains 6 cross-domain action recognition tasks. For fair comparison, all methods adopt the TRN~\cite{zhou2018temporal} as the backbone for video feature extraction, with the source model pre-trained on ImageNet~\cite{deng2009imagenet}. Following~\cite{liang2021source}, a Batch Normalization~\cite{ioffe2015batch} and an additional fully connected layer are inserted while weight normalization~\cite{salimans2016weight} is applied to the last fully connected layer. All experiments are implemented with PyTorch~\cite{paszke2019pytorch} library. \textit{More specifications on benchmark details and network implementation are provided in the Appendix.}

\subsection{Overall Results and Comparisons}
\label{section:exps:results}
We compare ATCoN with state-of-the-art SFDA approaches, as well as several competitive UDA/VUDA approaches. These include: SFDA~\cite{kim2021domain}, SHOT~\cite{liang2020we}, SHOT++~\cite{liang2021source}, MA~\cite{li2020model}, BAIT~\cite{yang2020bait} and CPGA~\cite{qiu2021cpga} which are designed for source-free adaptation; as well as DANN~\cite{ganin2015unsupervised}, MK-MMD~\cite{long2015learning} and TA\textsuperscript{3}N that are designed for UDA/VUDA scenario. We also report the results of the source-only model (TRN), which is obtained by applying the model trained with source data directly to the target data. We report the top-1 accuracy on the target domains, averaged on 5 runs with identical settings for each approach. Table \ref{table:4-1-sota-1} and Table \ref{table:4-2-sota-2} show the performance of our proposed ATCoN compared with the above methods in the three cross-domain action recognition benchmarks.

\begin{table}[!t]
\center
\caption{Results for SFVDA on UCF-HMDB\textsubscript{\textit{full}} and Sports-DA.}
\resizebox{.8\linewidth}{!}{\noindent
\begin{tabular}{c|c|ccc|ccccccc}
\hline
\hline
  \multirow{2}{*}{Methods} &
  \multirow{2}{*}{\parbox{0.12\linewidth}{\centering Source-free}} &
  \multicolumn{3}{c|}{\textbf{UCF-HMDB\textsubscript{\textit{full}}}} &
  \multicolumn{7}{c}{\textbf{Sports-DA}} \Tstrut\Bstrut\\
\cline{3-12}
& & U101$\to$H51 & H51$\to$U101 & Avg. & K600$\to$U101 & K600$\to$S1M & S1M$\to$U101 & S1M$\to$K600 & U101$\to$K600 & U101$\to$S1M & Avg.\\
\hline
TRN & -
& 72.78 & 72.15 & 72.47 & 86.41 & 66.95 & 85.31 & 71.05 & 49.29 & 43.32 & 67.06\\
\hline
DANN & \xmark
& 74.44 & 75.13 & 74.79 & 86.60 & 66.79 & 89.32 & 70.53 & 61.77 & 48.73 & 70.62\\
MK-MMD & \xmark
& 74.72 & 79.69 & 77.21 & 86.49 & 66.18 & 87.37 & 71.43 & 64.17 & \textbf{49.24} & 70.81\\ 
TA\textsuperscript{3}N & \xmark
& 78.14 & 84.83 & 81.49 & 88.24 & \textbf{70.56} & 83.32 & 75.54 & 57.51 & 46.37 & 70.26\\
\hline
SFDA & \cmark
& 69.86 & 74.98 & 72.42 & 86.10 & 60.02 & 85.37 & 68.04 & 55.75 & 43.58 & 66.48\\
SHOT & \cmark
& 74.44 & 74.43 & 74.44 & 91.19 & 64.95 & 88.84 & 72.02 & 53.93 & 43.58 & 69.09\\
SHOT++ & \cmark
& 71.11 & 68.13 & 69.62 & 90.01 & 63.11 & 88.01 & 70.34 & 44.75 & 40.95 & 66.20\\
MA & \cmark
& 74.45 & 67.36 & 70.91 & 91.04 & 65.95 & 87.84 & 71.88 & 60.75 & 39.41 & 69.48\\
BAIT & \cmark
& 75.33 & 76.36 & 75.85 & 92.27 & 66.61 & 88.33 & 72.85 & 57.25 & 44.67 & 70.33\\
CPGA & \cmark
& 75.82 & 68.16 & 71.99 & 89.42 & 66.26 & 86.49 & 72.55 & 55.22 & 44.53 & 69.08\\
\hline
\textbf{ATCoN} & \cmark
& \textbf{79.72} & \textbf{85.29} & \textbf{82.51} & \textbf{93.62} & \underline{69.70} & \textbf{90.64} & \textbf{75.99} & \textbf{65.24} & \underline{47.90} & \textbf{73.85}\Bstrut\\
\hline
\hline
\end{tabular}
}
\label{table:4-1-sota-1}
\end{table}

\begin{table}[!t]
\center
\caption{Results for SFVDA on Daily-DA.}
\resizebox{.92\linewidth}{!}{\noindent
\begin{tabular}{c|c|ccccccccccccc}
\hline
\hline
  \multirow{2}{*}{Methods} &
  \multirow{2}{*}{\parbox{0.12\linewidth}{\centering Source-free}} &
  \multicolumn{13}{c}{\textbf{Daily-DA}} \Tstrut\Bstrut\\
\cline{3-15}
& & K600$\to$A11 & K600$\to$H51 & K600$\to$MIT & MIT$\to$A11 & MIT$\to$H51 & MIT$\to$K600 & H51$\to$A11 & H51$\to$MIT & H51$\to$K600 & A11$\to$H51 & A11$\to$MIT & A11$\to$K600 & Avg. \Tstrut\\
\hline
TRN & -
& 20.87 & 36.66 & 29.00 & 22.11 & 43.75 & 53.10 & 13.81 & 22.00 & 37.10 & 17.20 & 14.75 & 24.38 & 27.89\\
\hline
DANN & \xmark
& 21.18 & 37.50 & 21.75 & 22.81 & 43.33 & \textbf{58.76} & 14.20 & 29.50 & 38.24 & 20.11 & \textbf{19.75} & 27.03 & 29.51\\
MK-MMD & \xmark
& \textbf{21.66} & 36.25 & 24.00 & 21.02 & \textbf{50.42} & 58.48 & \textbf{20.35} & 25.75 & 33.79 & 18.75 & 18.00 & 26.07 & 29.55\\ 
TA\textsuperscript{3}N & \xmark
& 19.87 & 37.67 & 31.53 & 21.57 & 43.01 & 55.47 & 14.38 & 25.71 & 38.39 & 14.92 & 15.56 & 23.42 & 28.49\\
\hline
SFDA & \cmark
& 12.57 & 44.95 & 27.50 & 15.96 & 35.19 & 49.23 & 13.08 & 24.25 & 24.86 & 16.29 & 13.25 & 25.22 & 25.19\\
SHOT & \cmark
& 12.03 & 44.58 & 29.50 & 15.28 & 36.67 & 51.04 & 13.58 & 24.25 & 21.24 & 17.08 & 14.00 & 24.35 & 25.30\\
SHOT++ & \cmark
& 12.57 & 40.83 & 28.75 & 14.90 & 41.67 & 46.34 & 15.98 & 22.25 & 33.10 & 15.42 & 12.50 & 21.76 & 24.42\\
MA & \cmark
& 12.76 & 45.82 & 30.00 & 17.75 & 37.36 & 53.54 & 12.90 & 25.00 & 22.19 & 16.67 & 15.25 & 24.29 & 26.13\\
BAIT & \cmark
& 12.69 & 45.73 & 30.00 & 16.93 & 39.64 & 53.00 & 13.65 & 25.50 & 21.17 & 15.70 & 14.50 & 25.52 & 26.17\\
CPGA & \cmark
& 13.06 & 46.02 & 30.75 & 18.08 & 39.21 & 55.09 & 13.14 & 26.25 & 25.54 & 19.19 & 16.50 & 26.72 & 26.46\\
\hline
\textbf{ATCoN} & \cmark
& \underline{17.21} & \textbf{48.25} & \textbf{32.50} & \textbf{27.23} & \underline{47.35} & \underline{57.66} & \underline{17.92} & \textbf{30.75} & \textbf{48.55} & \textbf{26.67} & \underline{17.25} & \textbf{31.05} & \textbf{33.53}\Bstrut\\
\hline
\hline
\end{tabular}
}
\label{table:4-2-sota-2}
\end{table}

The results in Table \ref{table:4-1-sota-1} and Table \ref{table:4-2-sota-2} show that the novel ATCoN achieves the best results among source-free methods on all 20 cross-domain tasks across the three cross-domain benchmarks, and outperforms previous source-free approaches considerably by noticeable margins. Notably, ATCoN exceeds all prior SFDA approaches designed for the image-based SFDA task (e.g., SHOT, MA, and CPGA) consistently by an average of more than $10\%$ relative improvements on mean accuracy over the second-best performances across all 18 cross-domain tasks. The consistent improvements empirically justify the effectiveness of learning temporal consistency for obtaining discriminative overall temporal features while attending to local temporal features with high source prediction confidence. Our proposed ATCoN even exceeds the performance of VUDA methods which are trained with accessible source data under 13 cross-domain tasks, while the mean accuracies of our method are consistently higher than all VUDA methods evaluated across the three benchmarks. This further validates the capability of ATCoN in constructing effective temporal features. 

Further, it could be observed that prior SFDA approaches could not tackle SFVDA well. Specifically, in 11 out of the 20 cross-domain tasks, more than half of the evaluated SFDA approaches result in performances inferior to that of the source-only model trained without any adaptation approaches. Prior SFDA approaches could only handle spatial features, while unable to obtain discriminative and transferrable temporal features, resulting in little or negative improvements compared to the source-only baseline. This further demonstrates the challenges faced when adapting video models under the source-free scenario. In particular, all tasks that involve ARID as the source or target domain in \textbf{Daily-DA} would lead to inferior results by prior SFDA approaches. This scenario could be further owed to the fact that videos in ARID are collected in adverse illumination with distinct statistical characteristics, leading to larger cross-domain gaps.

\begin{table}[!t]
\scriptsize
\center
\caption{Ablations studies of ATCoN on UCF-HMDB\textsubscript{\textit{full}}}
\subfloat[Components of temporal consistency\label{table:4-3-a-tc}]{
\resizebox{.45\textwidth}{!}{
\begin{tabular}[b]{c|cc}
    \hline
    \hline
    Methods & U101$\to$H51 & H51$\to$U101 \Tstrut\Bstrut\\
    \hline
    Source-only (TRN) & 72.78 & 72.15\Tstrut\\
    \textbf{ATCoN} & \textbf{79.72} & \textbf{85.29}\\
    \hline
    ATCoN-\textit{FC} & 77.78 & 83.36\Tstrut\\
    ATCoN-\textit{PC}\textsuperscript{$\dagger$} & 76.67 & 82.83\\
    ATCoN-\textit{PC} & 77.50 & 83.01\\
    ATCoN-\textit{TC} & 78.89 & 84.59\\
    \hline
    \hline
\end{tabular}
}
}
\qquad
\subfloat[\small Application of \textit{local relevance weight}\label{table:4-3-b-lwm}]{
\resizebox{.45\textwidth}{!}{
\renewcommand{\arraystretch}{1.2}%
\begin{tabular}[b]{c|cc}
    \hline
    \hline
    Methods & U101$\to$H51 & H51$\to$U101 \Tstrut\Bstrut\\
    \hline
    Source-only (TRN) & 72.78 & 72.15\Tstrut\\
    \textbf{ATCoN} & \textbf{79.72} & \textbf{85.29}\\
    \hline
    ATCoN-\textit{NA} & 78.33 & 83.89\Tstrut\\
    ATCoN-\textit{A@F} & 79.17 & 84.93\\
    ATCoN-\textit{A@P} & 78.61 & 84.41\\
    \hline
    \hline
\end{tabular}
}
}
\label{table:4-3-ablation}
\end{table}

\subsection{Ablation Studies and Feature Visualization}
\label{section:exps:ablation}
To dive deeper into the effectiveness of ATCoN and validate its architecture, we perform detailed ablation studies and feature visualization. The ablation studies investigate ATCoN from two perspectives: firstly, the components of temporal consistency; and secondly the application of \textit{local relevance weight} generated by \textit{LWM}. All ablation studies are conducted utilizing the \textbf{UCF-HMDB\textsubscript{\textit{full}}} dataset with 2 cross-domain action recognition tasks, while TRN is adopted as the feature extractor backbone.

\noindent\textbf{Temporal Consistency.}
We assess ATCoN against 4 variants to validate the design of the proposed temporal consistency loss $\mathcal{L}_{tc}$: \textbf{ATCoN-\textit{FC}}, where only the feature consistency is learnt; \textbf{ATCoN-\textit{PC}\textsuperscript{$\dagger$}} and \textbf{ATCoN-\textit{PC}}, where only the source prediction consistency is learnt, with the overall target prediction not included for \textbf{ATCoN-\textit{PC}\textsuperscript{$\dagger$}}; and finally \textbf{ATCoN-\textit{TC}}, where only the temporal consistency loss is learnt with both feature consistency and source prediction consistency. The above 4 variants would not apply both the IM loss and pseudo-label generation as proposed in Eq. \ref{eqn:method:im-t} and \ref{eqn:method:ce-tgt} during training, while the \textit{local relevance weight} in 
Sec. \ref{section:method:atcon:lwm}
is applied. Results in Table \ref{table:4-3-ablation}(a) demonstrate the efficacy of learning temporal consistency for constructing discriminative overall temporal features for tackling SFVDA. By learning either feature consistency or source prediction consistency, the network is able to outperform all prior SFDA approaches on both cross-domain tasks. Meanwhile, extending the source prediction consistency to the overall temporal feature further improves its efficacy. The superior performance of ATCoN-\textit{TC} justifies that learning feature consistency and source prediction consistency complements each other. 

Further, it could be observed that ATCoN performs slightly better than ATCoN-\textit{TC}, thanks to the inclusion of both IM loss and pseudo-labeling in training the full ATCoN. However, compared to the improvements towards the baseline model performance brought by learning temporal consistency, the performance gain by applying both IM loss and pseudo-labeling is marginal. The comparison empirically proves that the key towards ATCoN's success lies more in the learning of temporal consistency.

\noindent\textbf{Applying Local Relevance Weight.}
We propose the \textit{local relevance weight} $w_{lt}$ obtained from the \textit{LWM} which attends to the local temporal features with high confidence over their relevance to the source data distribution. To justify the necessity of the $w_{lt}$, we compare ATCoN against 3 variants: \textbf{ATCoN-\textit{NA}}, where the \textit{LWM} is not inserted thus $w_{lt}$ is not obtained at all; \textbf{ATCoN-\textit{A@F}}, where $w_{lt}$ is only applied for obtaining the overall temporal feature $\mathbf{t}_{T}^{\prime}$; and \textbf{ATCoN-\textit{A@P}}, where $w_{lt}$ is only applied to obtain the weighted local source prediction ${p_{lt,T}^{(r)}}^{\prime}$. Both the IM loss and pseudo-label generation are adopted during the training of the three aforementioned variants. As illustrated in Table \ref{table:4-3-ablation}(b), applying \textit{local relevance weight} brings consistent improvements wherever it has been applied, which justifies the necessity for such a weight. By employing the \textit{local relevance weight} $w_{lt}$, ATCoN is able to obtain more discriminative temporal features. While $w_{lt}$ bring further improvements on network performance, it should be noted that the improvement is relatively marginal compared to that brought by learning temporal consistency, which indicates that the proposed temporal consistency plays a more vital role in tackling SFVDA effectively.

\begin{figure}[!t]
\begin{center}
   \includegraphics[width=.9\linewidth]{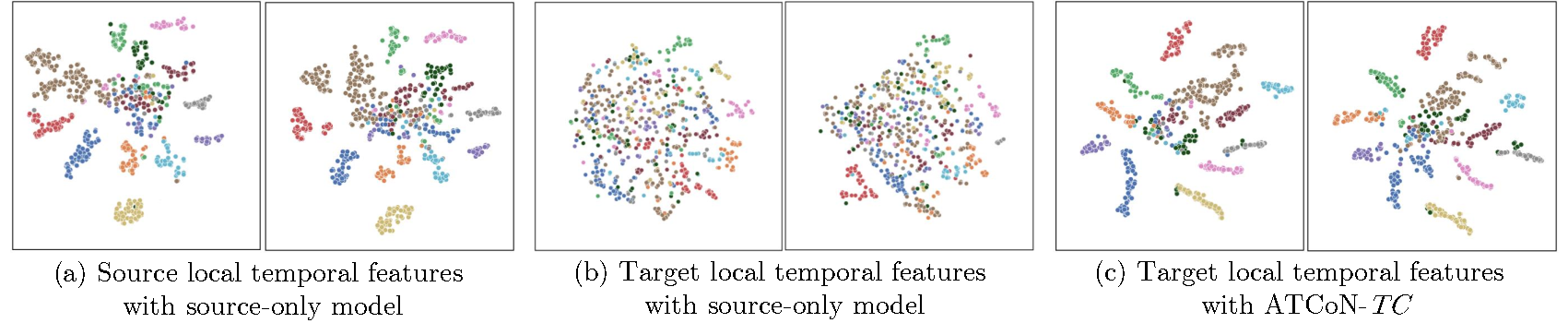}
\end{center}
\caption{t-SNE visualizations of local temporal features with class information. Different colors represent different classes.}
\label{figure:3-2-lt-visual}
\end{figure}

\begin{figure}[!t]
    \center
    \subfloat[Source-Only\label{figure:4-3-a-trn}]{
    \includegraphics[width=.18\textwidth, frame]{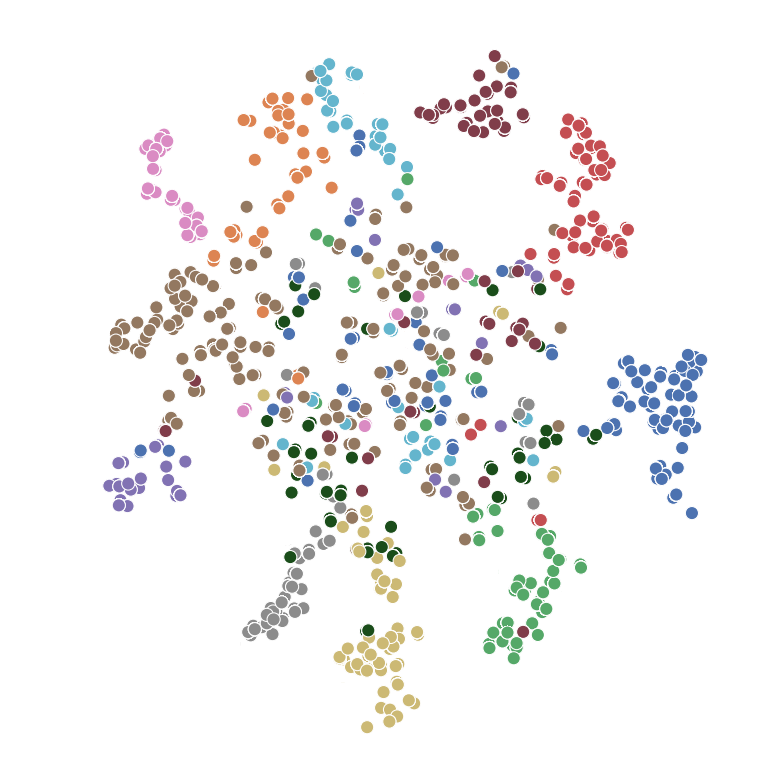}}
    \subfloat[CPGA\label{figure:4-3-b-cpga}]{
    \includegraphics[width=.18\textwidth, frame]{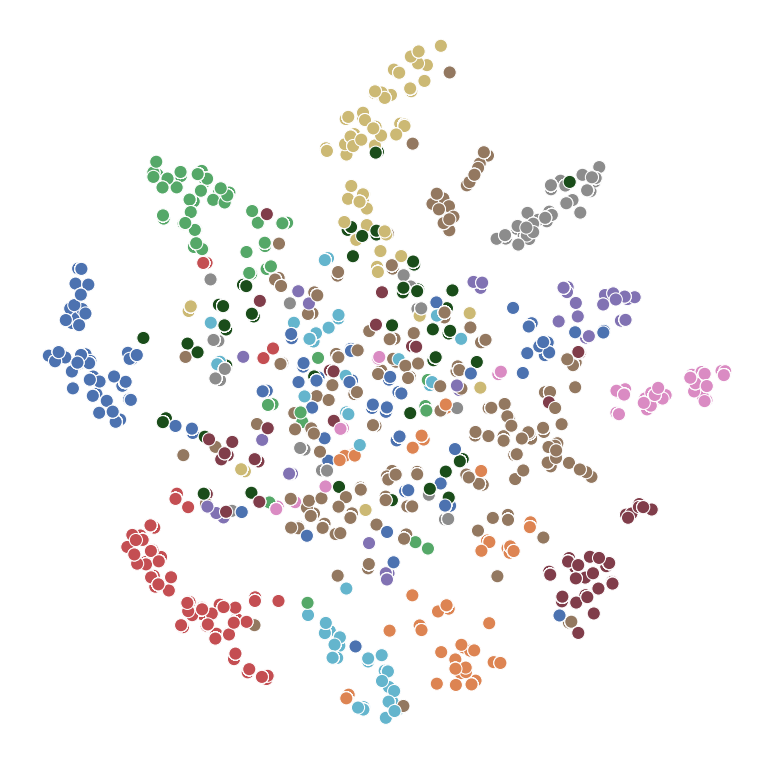}}
    \subfloat[SHOT\label{figure:4-3-c-shot}]{
    \includegraphics[width=.18\textwidth, frame]{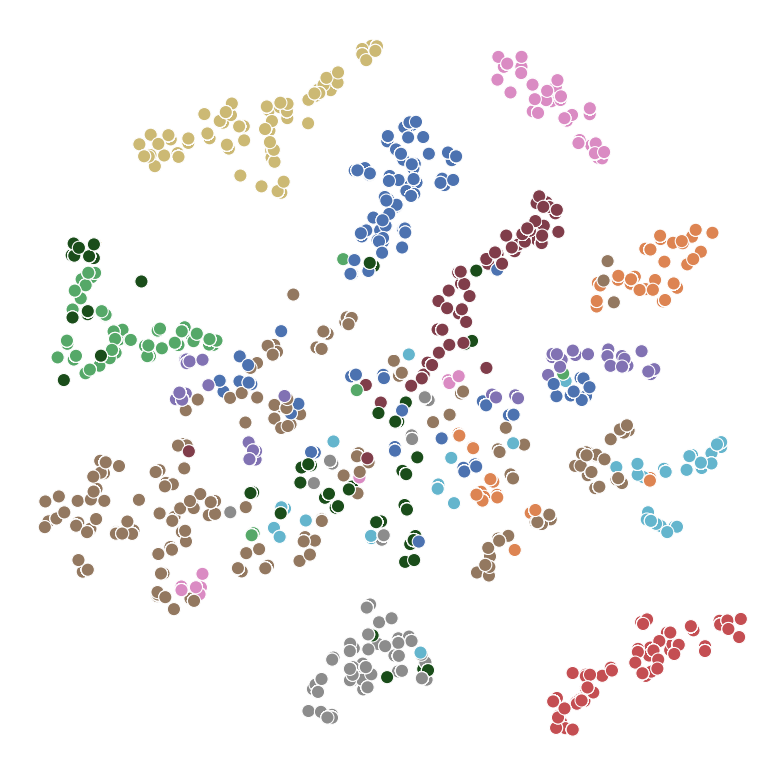}}
    \subfloat[ATCoN\label{figure:4-3-a-atcon}]{
    \includegraphics[width=.18\textwidth, frame]{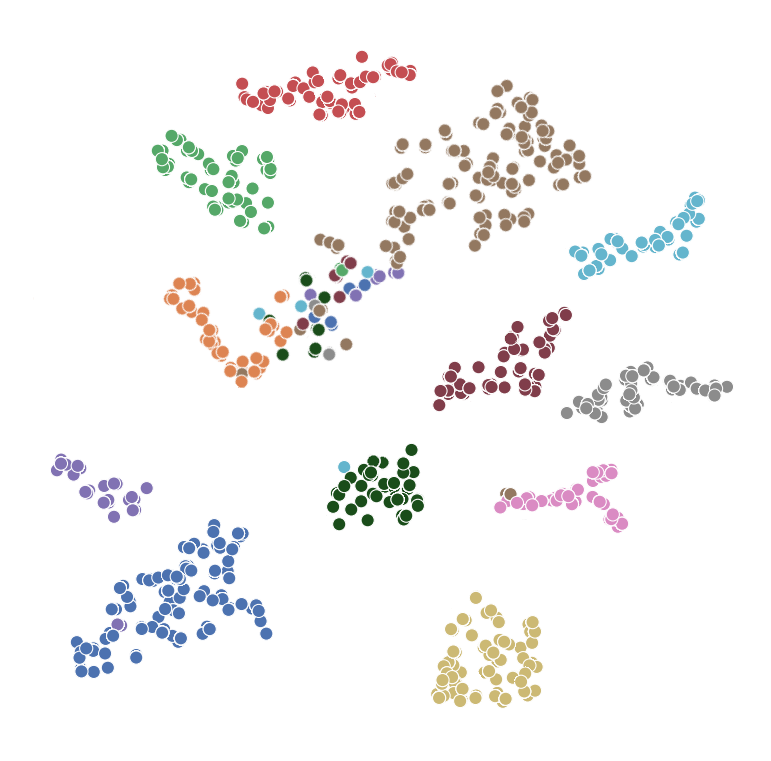}}
    \caption{Visualization of features extracted by the (a) source-only model, (b) CPGA, (c) SHOT, and (d) ATCoN with class information. Different classes are marked by different colors.}
    \label{figure:4-3-tsne}
\end{figure}

\noindent\textbf{Feature Visualization.}
To further understand the characteristics of ATCoN, we plot the t-SNE embeddings~\cite{van2008visualizing} of the features extracted. Specifically, we first prove our \textit{cross-temporal hypothesis} by visualizing local temporal features learned by the source-only model on the source data and the target data, and local temporal features learned by ATCoN-\textit{TC} for the H51$\to$U101 task, as presented in Fig. \ref{figure:3-2-lt-visual}. The local temporal features of the source data share similar distribution patterns, which confirms that they are both discriminative and consistent, with similar semantic information embedded. Meanwhile, the data distribution patterns of target data with the source model are inconsistent. In comparison, by learning temporal consistency, ATCoN-\textit{TC} is able to extract discriminative and relatively consistent local temporal features, satisfying the \textit{cross-temporal hypothesis}. This implies that learning temporal consistency enables the learning of source-like representations for target data, and therefore is effective in aligning target data to source data distribution.

We further plot the t-SNE embeddings of the overall temporal features learnt by ATCoN, CPGA, and SHOT for the H51$\to$U101 task with class information in the target domain. The results are presented in Fig. \ref{figure:4-3-tsne}, where we can clearly observe that the features learned by ATCoN are much more clustered than those learned by other networks. This verifies that features learned by ATCoN are of higher discriminability, resulting in better SFVDA performance. In contrast, features learned by CPGA are even less clustered and discriminative than those learned by the source-only backbone, which corresponds to its inferior performance over the backbone in this task. The above observation implies the superiority of our ATCoN in tackling SFVDA while reflecting the challenges faced by prior SFDA approaches in tackling SFVDA.

\section{Conclusion}
\label{section:concl}

In this work, we pioneer in formulating the challenging yet realistic Source-Free Video Domain Adaptation (SFVDA) problem, which addresses data-privacy issues in videos. We proposed a novel ATCoN to tackle SFVDA effectively. With source video data inaccessible, ATCoN tackles SFVDA via obtaining effective and discriminative overall temporal features satisfying the \textit{cross-temporal hypothesis}, achieved by learning temporal consistency, guaranteed by both feature consistency and source prediction consistency. ATCoN further aims to align target data to the source distribution through attending to local temporal features with higher source prediction confidence. Extensive experiments and detailed ablation studies across multiple cross-domain action recognition benchmarks validate the superiority of our proposed ATCoN in tackling SFVDA.



%
%
\bibliographystyle{splncs04}
\bibliography{eccv}
\end{document}